\documentclass[10pt,twocolumn,letterpaper]{article}

\usepackage{wacv}
\usepackage{times}
\usepackage{epsfig}
\usepackage{graphicx}
\usepackage{amsmath}
\usepackage{amssymb}
\usepackage{amsfonts}
\usepackage{amsthm}
\usepackage{graphicx}
\usepackage{dsfont}
\usepackage{listings}
\usepackage{algorithm}
\usepackage{algpseudocode}
\usepackage{multirow}
\usepackage{xcolor}

\newtheorem{theorem}{Theorem}
\newtheorem{lemma}[theorem]{Lemma}
\usepackage{microtype}
\usepackage[caption=false]{subfig}

\wacvfinalcopy
\providecommand{\doarxiv}{false}
\usepackage{ifthen}
\newboolean{isarxiv}
\setboolean{isarxiv}{\doarxiv}
\ifthenelse{\boolean{isarxiv}}{%
\newcommand{\arxiv}[1]{#1}
\newcommand{\notarxiv}[1]{}
}{
\newcommand{\arxiv}[1]{}
\newcommand{\notarxiv}[1]{#1}
}

\newcommand{\arxivalt}[2]{\ifthenelse{\boolean{isarxiv}}{#1}{#2}}
\newcommand{\arxivaltr}[2]{\ifthenelse{\boolean{isarxiv}}{#2}{#1}}
\newcommand{\narxiv}[1]{\notarxiv{#1}}
\def\wacvPaperID{647} % *** Enter the wacv Paper ID here

\ifwacvfinal\fi
\begin{document}

%%%%%%%%% TITLE
\title{A Framework towards Domain Specific Video Summarization}

% Authors at the same institution
%\author{First Author \hspace{2cm} Second Author \\
\author{Vishal Kaushal \\ IIT Bombay \\ {\tt\small vkaushal@cse.iitb.ac.in}
\and Sandeep Subramanian \\ IIT Bombay\\ {\tt\small sandeeps94@cse.iitb.ac.in}
\and Suraj Kothawade \\ IIT Bombay\\{\tt\small surajkothawade@cse.iitb.ac.in} \and  Rishabh Iyer \\ Microsoft Corporation \\ {\tt\small rishi@microsoft.com} \and Ganesh Ramakrishnan \\ IIT Bombay \\ {\tt\small ganesh@cse.iitb.ac.in}}

%\affil{Indian Institute of Technology Bombay$^1$, Microsoft Corporation$^2$}

% Authors at different institutions
%\author{Rishabh Iyer \\
%{\tt\small rishi@microsoft.com}
%}
\maketitle
\ifwacvfinal\thispagestyle{empty}\fi

%%%%%%%%% ABSTRACT

\begin{abstract}
In the light of exponentially increasing video content, video summarization has attracted a lot of attention recently due to its ability to optimize time and storage. Characteristics of a good summary of a video depend on the particular domain under question. We propose a novel framework for domain specific video summarization. Given a video of a particular domain, our system can produce a summary based on what is important for that domain in addition to possessing other desired characteristics like representativeness, coverage, diversity etc. as suitable to that domain. Past related work has focused either on using supervised approaches for ranking the snippets to produce summary or on using unsupervised approaches of generating the summary as a subset of snippets with the above characteristics. We look at the joint problem of learning domain specific importance of segments as well as the desired summary characteristic for that domain. Our studies show that the more efficient way of incorporating domain specific relevances into a summary is by obtaining ratings of shots as opposed to binary inclusion/exclusion information. We also argue that ratings can be seen as unified representation of all possible ground truth summaries of a video, taking us one step closer in dealing with challenges associated with multiple ground truth summaries of a video. We also propose a novel evaluation measure which is more naturally suited in assessing the quality of video summary for the task at hand than F1 like measures. It leverages the ratings information and is richer in appropriately modeling desirable and undesirable characteristics of a summary. Lastly, we release a gold standard dataset for furthering research in domain specific video summarization, which to our knowledge is the first dataset with long videos across several domains with rating annotations. We conduct extensive experiments to demonstrate the benefits of our proposed solution.\looseness-1
\end{abstract}
%%%%%%%%% BODY TEXT
\section{Introduction}
With the explosion of video data, automatic analysis of videos is increasingly becoming important. Examples of such videos include user videos, sports videos, TV videos or CCTV footages or for that matter videos coming from any other source. One of the popular requirements in this context is the ability to automatically summarize videos. Video summarization finds its uses in a wide variety of applications ranging from security and surveillance to compliance and quality monitoring to user applications aimed at saving time and storage. Broadly speaking, in terms of the kind of video summaries produced video summarization can be of two types - compositional video summarization, which aims at producing spatio-temporal synopsis \cite{pritch2007webcam,rav2006making,pritch2009clustered,pritch2008nonchronological} and extractive video summarization, which aims at selecting key frames (also called key frame extraction, static story boards or static video summarization, eg. \cite{de2011vsumm}) or shots (also called dynamic video summarization or dynamic video skimming, eg. \cite{GygliECCV14}). Past work has tried to address the problem of video summarization using unsupervised as well as supervised techniques. Early unsupervised techniques used attention models \cite{ma2005generic}, MMR \cite{li2010multi}, clustering \cite{de2011vsumm,mahmoud2013unsupervised} etc. and more recently auto encoder based techniques \cite{yang2015unsupervised} and LSTM based techniques \cite{mahasseni2017unsupervised} have been used. Use of supervised techniques began with forms of indirect supervision from external sources of information like related images \cite{khosla2013large}, similar videos \cite{chu2015video} or title of videos \cite{song2015tvsum}. Gong et al's work \cite{gong2014diverse} was the first work on video summarization which used direct high level supervision in form of user annotations. This was followed by Gygli et al's work \cite{gygli2015video}. Sharghi, Gong et al took this forward by incorporating a notion of user query in the produced video summaries thereby involving the user in the summary generation process \cite{sharghi2016query,sharghi2017query}. Theirs became the first work on query focussed video summarization. Then came several deep learning based video summarization techniques. For example, Zhang et al. \cite{zhang2016video} used LSTMs \cite{hochreiter1997long} to model long range dependencies among video key frames to produce summaries. Emphasis was given to sequential structures in videos and their modeling. Mahasseni et al.\cite{mahasseni2017unsupervised} used LSTM networks in an adversarial setting to generate summaries. Both these methods were domain agnostic. Some researchers viewed video summarization as a subset selection problem \cite{zhang2016summary,elhamifar2017online,sharghi2016query,sharghi2017query}. Another key approach was to combine the elements of both deep learning and subset selection in summary generation. Gong et al. achieved this using their seqDPP architecture \cite{gong2014diverse} where an LSTM network was coupled with a DPP (Determinantal Point Process) \cite{kulesza2012determinantal}.

In this work we address the challenge of domain specific dynamic video summarization. A good video summary should be a meaningful short abstract of a long video, it must contain \emph{important} events, it should exhibit some continuity and it should be free from redundancy. However, the notion of importance varies with domain. For example a "six" or a "wicket" could be considered important in cricket videos while "entry of birthday girl" and "cutting of cake" would be considered important in birthday videos. Further the characteristics of a good summary like representativeness, coverage or diversity also depends on the domain. For example, while for surveillance videos a good summary should contain outliers, for a user video coverage and representativeness become more important. We hereby propose a novel domain specific video summarization framework which automatically learns to produce summaries that possess desired characteristics suitable for that domain. Past work on video summarization has focused either on using supervised approaches for ranking the snippets/segments thereby producing a video summary (eg. \cite{potapov2014category,ma2005generic}) or on unsupervised approaches of generating the video summary as a subset of snippets/segments with desired characteristics of representativeness, diversity, coverage, etc. (eg. \cite{li2010multi,de2011vsumm}). Some work has also focused on learning the relative importance of uniformity, interestingness and representativeness for different domains (eg. \cite{gygli2015video,sahoo2017unified}). None of these works, however, have looked at the joint problem of generating domain specific summaries by automatically learning the concepts that are deemed important for a domain, together with having the desired summary characteristics like diversity, coverage etc. for that domain. Building upon the max margin structured learning framework in \cite{gygli2015video} we learn a mixture of modular and submodular terms. Modular terms help to capture shots more important to the domain under consideration and submodular terms help to capture characteristics of summary important to that domain. The different weights learnt for different components indicate the varying notion of importance from domain to domain. 

Further, having many possible ground truth summaries has been posed as one of the challenges in video summarization~\cite{truong2007video}. Multiple ground truths are due to the difference in perspectives and the fact that different visual content can have the same semantic meaning. However, one reason is also a lack of any other information about the video. Dealing with domain specific videos, however, allows us to define a notion of importance ratings that are unique and unambiguous for that domain. Such ratings can be seen as a unified representation of all possible ground truth summaries of a video. We establish the importance of ratings in the training dataset in producing good summaries as against binary inclusion/exclusion information. Instead of getting ground truth user summaries from annotators we rather ask them to provide ratings for segments in the entire video. The framework learns to generate more accurate summaries when it is provided more supervision this way. We thus establish that the more efficient way of incorporating domain specific relevances into a summary is to provide a supervision in form of ratings as against multiple ground truths.

We also define a new evaluation measure which is more naturally suited for this task than other standard measures used in literature like F1 score. Our measure evaluates video summaries considering the ratings and not just binary inclusion/exclusion information. It also evaluates summaries not only with respect to what they \emph{should} contain but also with respect to what they \emph{should not} contain and the degree of diversity.

As a part of this work we will also release a gold standard dataset for furthering research in domain specific video summarization. To the best of our knowledge, ours is the first dataset with long videos across several domains with rating annotations.

\section{Related Work and Our Contributions}

\subsection{Domain specific video summarization}

One of the earliest works on domain specific video summarization was by Potapov et al.\cite{potapov2014category} in 2014. They were one of the first to realize the importance of building separate models for summarization for distinct categories of videos. They used an SVM\cite{hearst1998support} classifier conditioned on video category to produce summaries. The SVM learns to score the segments according to their importance to the domain. The segments having higher scores are then selected greedily and put in temporal order to create the final summary. Sun et al. \cite{sun2014ranking} analyzed edited videos of a particular domain as an indicator of highlights of that domain. After finding pairs of raw and corresponding edited videos, they obtain pair-wise ranking constraints to train their model. Zhang et al. \cite{zhang2016summary} use supervision in the form of human-created summaries to perform automatic keyframe-based video summarization. Their main idea is to nonparametrically transfer summary structures of a particular domain of videos from annotated videos to unseen test videos. By learning a joint model to understand what snippets and what characteristics are important for a domain, we use a more principled approach with a form of supervision which is more efficient.

\subsection{Submodular functions for video summarization}

Video summarization can be viewed as a subset selection problem subject to certain constraints. Given a set $V = \{1, 2, 3, \cdots, n\}$ of items which we also call the \emph{Ground Set}, define a utility function (set function) $f:2^V \rightarrow \mathbf{R}$, which measures how good a subset $X \subseteq V$ is. Let $c :2^V \rightarrow \mathbf{R}$ be a cost function, which describes the cost of the set (for example, the size of the subset). Often the cost $c$ is budget constrained (for example, a fixed set summary) and a natural formulation of this is the following problem:
\begin{align}
\max\{f(X) \mbox{ such that } c(X) \leq b\}
\end{align}

The goal is then to have a subset $X$ which maximizes $f$ while simultaneously minimizing the cost function $c$. It is easy to see that maximizing a generic set function becomes computationally infeasible as $V$ grows. 

A special class of set functions, called submodular functions ~\cite{nemhauser1978analysis}, however, makes this optimization easy. A function $f : 2^V \rightarrow R$ is submodular if for every
$A \subseteq B \subseteq V$ and $e \in V$ and $e \notin B$ it holds that
\begin{align}
f(\{e\} \cup A) - f(A) \geq f(\{e\} \cup B) - f(B)
\end{align}
Likewise,a function $f : 2^V \rightarrow R$ is supermodular if for every
$A \subseteq B \subseteq V$ and $e \in V$ and $e \notin B$ it holds that
\begin{align}
f(\{e\} \cup A) - f(A) \leq f(\{e\} \cup B) - f(B)
\end{align}
Submodular functions exhibit a property that intuitively formalizes the idea of ``diminishing returns''. That is, adding some instance $x$ to the set $A$ provides more gain in terms of the target function than adding $x$ to a larger set $A'$, where $A \subseteq A'$.  Informally, since $A'$ is a superset of $A$ and already contains more information, adding $x$ will not help as much. Using a greedy algorithm to optimize a submodular function (for selecting a subset) gives a lower-bound performance guarantee of around 63\% of optimal ~\cite{nemhauser1978analysis} to the above problem, and in practice these greedy solutions are often within 98\% of optimal ~\cite{krause2008optimizing}. This makes it advantageous to formulate (or approximate) the objective function for data selection as a submodular function. 

This concept has been used in document summarization \cite{lin2012learning} and in image collection summarization \cite{tschiatschek2014learning}. More recently, Elhamifar et al. \cite{elhamifar2017online} used submodular optimization for online video summarization by performing incremental subset selection.  One of the first attempts to summarize videos using a submodular mixture of objectives was by Gygli et al.~\cite{gygli2015video}. They, however, did not distinguish between various domains of videos and had a specially crafted video frame interestingness model which played a significant role in the summaries produced. Building upon the approach in~\cite{gygli2015video} we learn a mixture for a domain to produce summary specific to that domain. However, in our work, domain specific importance of snippets/shots is not predicted separately using another model. Rather weighted features are directly used in the mixture as modular terms. These modular components capture the shot level domain importance while the other submodular and supermodular components in the mixture correspond to different desired characteristics of the summary like diversity and coverage.

\subsection{Evaluation measures}

Different measures have been reported in literature for the purpose of accurately evaluating the quality of the video summary produced. VIPER \cite{doermann2000tools} addresses the problem by defining a specific ground truth format which makes it easy to evaluate a candidate summary. SUPERSEIV \cite{huang2004automatic} is an unsupervised technique to evaluate video summarization algorithms that perform frame ranking. VERT \cite{li2010vert} was inspired by BLEU in machine translation and ROUGE in text summarization. More recently approaches by Yeung et al. \cite{yeung2014videoset} and Plummer et al. \cite{plummer2017enhancing} also used text based evaluation methods. De Avila et al. \cite{de2011vsumm} propose a method of evaluation which considers several ground truth summaries. Others, like \cite{khosla2013large,GygliECCV14,elhamifar2017online} and \cite{xu2015gaze} more directly use precision and recall type measures. Kannappan et al. \cite{kannappan2016pertinent} propose an approach which is only for static video summaries. The evaluation measure proposed by Potapov et al. \cite{potapov2014category} is capable of evaluating a summary only against one ground truth. As annotations are done by several users producing several ground truth summaries, they evaluate those annotations against each other to form some kind of an upper bound on performance. Approaches like \cite{zhao2014quasi,conf/cvpr/SongVSJ15,gygli2015video} and \cite{zhang2016video} combine several ground truths into one before using them for evaluation. This comes at the cost of losing individual opinion. In search for a measure which would work directly on ratings (which is a potential generator of multiple ground truths) and having certain other desired characteristics (as enumerated in the corresponding section below) we developed our own evaluation measure. 

\subsection{Datasets for video summarization}

Different researchers in the past have released different datasets for the purpose of video summarization. Examples include The Video Summarization (SumMe) dataset \cite{GygliECCV14}, MED Summaries dataset \cite{potapov2014category} and Title-based Video Summarization (TVSum) dataset\cite{conf/cvpr/SongVSJ15}. However, none of these existing datasets were found suitable for our work for following reasons. Firstly, we aim to summarize videos across a large number of domains like surveillance, sports, user etc. in a single framework. For that purpose we need a wide variety of videos summarized uniformly. The various datasets only provide certain subsets of types of videos and they use vastly different methods to annotate those videos. So, it was essential that we had a uniformly annotated, diverse set of videos from diverse domains. Secondly, we wanted to test our method on long videos, as the true benefit of a video summary in real world applications is seen only with respect to long videos. Thirdly, we wanted the annotations to not only capture what is important, but also what is \emph{not} important and what is repetitive. Identifying segments which are relatively long and contain repetitive information  (for example, scene of spectators clapping for 5 minutes in a cricket video) and retaining only a fraction of them to be included in the summary, is essential to having a good quality summary. 

\subsection{Our Contributions}
In the following, we summarize the main contributions of this paper.
\begin{itemize}
\item We address the problem of Domain specific video summarization, by jointly ranking the most important portions of the video for that domain (for example, a goal in Soccer), while simultaneously capturing diverse and representative shots. We do this by training a joint mixture model with features which capture domain importance along with diversity models.
\item We introduce a novel evaluation criteria which captures these aspects of a summary, and also introduce a large dataset for domain specific summarization. Our dataset comprises of several long videos for different domains (surveillance, personal videos, sports etc.) and to our knowledge is the first domain specific video summarization dataset with long videos.
\item We then empirically demonstrate various interesting insights. a) We first show that by jointly modeling diversity, relevance and importance, we can learn substantially superior summaries on all domains compared to just learning any one of these aspects. b) We next show that by learning on the same domain, we can obtain superior results than using learnt mixtures from other domains, thus proving the benefit for domain specific video summarization. c) We then look at the top components learnt for different domains and show how those individual components perform best for that domain if considered in isolation. Moreover, we observe how intuitively it makes sense that these components are important. For example, in surveillance, we see that diversity functions tend to have high ranking compared to other models (as finding outliers is more important there), while in personal videos (like birthday), we see that representation is important. We moreover also look at the highest ranking snippets based on these components and show how they capture the most important aspects of that domain (like a goal in soccer).
\end{itemize}
The major contribution of this work is that this is the first systematic study of domain specific video summarization on large videos and we provide several insights into the role of different summarization models for this problem.
\section{Methodology}

We begin by creating a training dataset comprising of videos from several categories annotated with ratings information. Our method works on ratings and hence better deals with issue of multiple ground truths. Building upon the approach in \cite{gygli2015video}, we create a mixture, but our mixture contains modular terms (to capture the domain specific importance of snippets) and submodular terms (for imparting certain desired characteristics to the summary). For each training video of a domain, the components of the mixture are instantiated and the weights of the complete mixture for that domain are learnt using max margin learning framework. After the training phase, for any given test video of that domain, the weighted mixture is then maximized to produce the desired summary video. Below we describe details of every step in the above methodology.

\subsection{Training Data}

In this work we focus on videos from five different domains - birthday, cricket, soccer, entry exit and office. The latter two are surveillance videos taken from CCTV cameras installed at various entry/exit locations and offices respectively. We have collected birthday, cricket and soccer videos from internet (existing published datasets / youtube). Due to privacy reasons and to be able to experiment with presence/absence of abnormal events in the surveillance videos, we have collected surveillance videos from our own setup of surveillance cameras. \arxiv{Table \ref{tab:table1} shows distribution of number and duration of videos across these categories.

\begin{table}[h!]
  \begin{center}
    \caption{Distribution of videos across categories}
    \label{tab:table1}
    \begin{tabular}{l|l|l} 
      \textbf{Category} & \textbf{Number of Videos} & \textbf{Duration in mins}\\
      \hline
      Cricket & 7 & 276\\
      Birthday & 9 & 136\\
      Soccer & 11 & 609\\
      Entry Exit & 21 & 306\\
      Office & 33 & 687\\
    \end{tabular}
  \end{center}
\end{table}}
\narxiv{We have 7 Cricket videos (of 276 mins), 9 of Birthday (136 minutes total length) and 21 of Entry Exit (306 minutes).}

Next, for each domain, we go over every video and first prepare a table of different kinds of scenes that occur across different videos in that category and using domain knowledge we assign ratings to those scenes. Negative ratings are assigned to segments which \emph{must not} be included in the summary. Since the ratings are relative (for example, a 2 rated scene is supposed to be more important than a 1 rated scene but less important than a 3 rated scene) it was necessary to gather information about all scenes before starting to rate the scenes in the specific videos. Also going through the extra step of creating such a scenes document for each category enabled us to come up with a consistent philosophy of ratings and consequently a very high inter annotator correlation. Using this scenes table as annotation guidelines for each category, the annotators then annotated the videos in each category. For each video, three different annotators were independently asked to identify the segments based on the scenes and label them. Segments that are long and contain repetitive content are explicitly marked repetitive in addition to their rating. For the purpose of annotating, we customized a tool called oTranscribe \cite{oTranscribe} to make the annotation task easy and to produce the desired annotation JSON. The oTranscribe interface was cleaned up and a lot more keyboard shortcuts were added in for ease of annotation. Shortcuts were added in to mark the beginning and ends of segments, to rate segments, to give a short description of the segment, to mark a segment repetitive and to skip to the end and beginning of the previous segment. Finally, hooks were added in to output the annotation as a JSON file. The annotations from different annotators were combined based on majority agreement. As a sanity check, we visually verified the annotations thus produced by looking at the annotated videos which were produced by overlaying the labels on top of the original videos.

\subsection{Learning framework}

The task of video summarization is posed as a discrete optimization problem for finding the best subset representing the summary. Given a video $V$ we split it into a snippets $v_i$ of fixed length. Now we have a set $Y_v$ of all snippets in the video. Our problem reduces to picking $y$ $\subset$ $Y_v$ such that $|y| \leq k$ that maximizes our objective.\looseness-1
\begin{align} \label{max}
y^* = \operatorname*{argmax}_{y\subseteq Y_v, |y| \leq k} o(x_v, y)
\end{align}
$y^*$ is the predicted summary, $x_v$ the feature representation of the video snippets and $o(x_v, y)$ is the weighted mixture of components each capturing some aspect of the domain. Different weights are learnt for different domains.\looseness-1
\begin{align}
o(x_v, y) = w^Tf(x_v, y)
\end{align}
where $f(x_v, y) = [f_1(x_v, y),..., f_n(x_v, y)]$ and $f_i(x_v, y)$ are the various modular, submodular and supermodular components. Given $N$ pairs of a video and a reference summary $(V, y_{gt})$, we learn the weight vector $w$ by optimizing the following large-margin \cite{taskar2005learning} formulation:\looseness-1
\begin{align} \label{min}
\min\limits_{w \geq 0} \frac{1}{N} \sum_{n=1}^{N} L_n(w) + \frac{\lambda_1}{2}||w_1||^2 + \frac{\lambda_2}{2}||w_2||^2
\end{align}
where $L_n(w)$ is the generalized hinge loss of training example $n$ and $w_1$ and $w_2$ are the weight vectors for the modular terms and the submodular terms respectively.\looseness-1
\begin{align} \label{loss-aug}
L_n(w) = \max\limits_{y \subseteq Y_v^n} (w^T f(x_v^n, y) + l_n(y)) - w^T f(x_v^n, y_{gt}^n)
\end{align}
This objective is chosen so that each human reference annotation scores higher than any other summary by some margin. For training example $n$, the margin we chose is denoted by $l_n(y)$. We use $1 - normalizedScore(y)$ as margin where normalized score is computed using min-max normalization of the score generated by our evaluation measure, given the ratings, as described below.

\subsection{Components of the mixture}

Our mixture contains several hand picked components. Every component serves to impart certain characteristics to the optimal subset (the predicted summary). 

\textbf{Set Cover:} For a subset $X$ being scored, the set cover is defined as $f_{sc}(X) = \sum_{u \in U} min\{m_u(X), 1\}$  $u$ is a concept belonging to a set of all concepts $U$, $m_u(X) = \sum_{x \in X} w_{xu}$ and $w_{xu}$ is the weight of coverage of concept $u$ by element $x$. This component governs the coverage aspect of the candidate summary and is monotone submodular.\looseness-1

\textbf{Probabilistic Set Cover:} This variant of the set cover function is defined as $f_{psc}(X) = \sum_{u \in U} (1 - \prod_{x \in X} (1 - p_{xu}))$ where $p_{xu}$ is the probability with which concept $u$ is covered by element $x$. Similar to the set cover function, this function governs the coverage aspect of the candidate summary, viewed stochastically and is also monotone submodular.\looseness-1

\textbf{Facility Location:} The facility location function is defined as  $f_{fl}(X) = \sum_{v \in V} \max_{x \in X} sim(v,x)$ where $v$ is an element from the ground set $V$ and $sim(v, x)$ measures the similarity between element v and element x. Facility Location governs the representativeness aspect of the candidate summaries and is monotone submodular.\looseness-1

\textbf{Saturated Coverage} Saturated Coverage is $f_{satc}(X) = \sum_{v \in V} min\{m_v(X), c\}$ where $m_v(X) = \sum_{x \in X} sim(v, x)$ measures the relevance of set $X$ to item $v \in V$ and $c$ is a saturation hyper parameter that controls the level of coverage for each item $v$ by the set $X$. Saturated Coverage is similar to Facility Location except for the fact that for every category, instead of taking a single representative, it allows for taking potentially multiple representatives. Saturated Coverage is also monotone submodular.\looseness-1

\textbf{Generalized Graph-Cut} Generalized Graph Cut is $f_{gc}(X) = \sum_{i \in V, j \in X} sim(i, j) - \lambda \sum_{i, j \in X} sim(i, j)$
Similar to above two functions, Generalized Graph Cut also models representation. When $\lambda$ becomes large, it also tries to model diversity in the subset. $\lambda$ governs the tradeoff between representation and diversity. For $\lambda < 0.5$ it is monotone submodular. For $\lambda > 0.5$ it is non-monotone submodular. \looseness-1

\textbf{Disparity-min:} Denoting the distance measure between snippet/shot $i$ and $j$ by $d_{ij}$, disparity-min is defined as a set function $f_{disp}(X) = \min_{i, j \in X, i \neq j} d_{ij}$. It is easy to see that maximizing this function involves obtaining a subset with maximal minimum pairwise distance, thereby ensuring a diverse subset of snippets or shots. In principle this is similar to determinantal point processes (DPP), but DPP becomes computationally expensive at inference time. This function, though not submodular, can be efficiently optimized via a greedy algorithm.

\textbf{Continuity:} We work on a set of 2 second snippets as ground set. A summary (subset) could thus may not look continuous enough to give good viewing experience. Thus we add this continuity term in the mixture which would give more score when nearby snippets are chosen - this would ensure a visually more coherent and appealing summary. Essentially it is modeled as a redundancy function (this function is super-modular) within a shot as follows: $f_{cont}(X) = \sum_{s \in S} \sum_{x, x' \in s \cap X} w_{x, x'}$ where $S$ is the set of shots as a result of a shot boundary detection algorithm and $w_{x,x'}$ is the similarity between two snippets which can be defined as how close they are to each other based on their index. That is, the features used here are the indices of the snippets.

\textbf{Modular components:}
We use weighted features of snippets (described in the next section) as modular terms in the mixture.\\

\subsection{Features used for instantiating the components}
Let the video $\mathcal{V}$ be a set of frames $f_i, i = \{0, 1, 2, \cdots, n\}$. Let us define the ground set $V = \{S_0, S_1, S_2, \cdots, S_k\}$ as a set of \emph{snippets} where each snippet is 2 seconds long. A snippet  for a video with frame rate $r$ would thus contain $2r$ consecutive frames. Feature vectors are calculated for each snippet independently by averaging the feature vectors of the frames/images in that snippet. For  \narxiv{Different components of the mixture (as above) are instantiated for each video using the features such as VGG~\cite{simonyan2014very}, GoogleNet~\cite{szegedy2015going}, YOLO entities and features from Pascal VOC and COCO~\cite{redmon2016you} and Color Histogram features. A comprehensive list of the features used and how they are computed can be found in the extended version of this paper in the supplementary material}. \arxiv{Different components of the mixture (as above) are instantiated for each video using the features mentioned below:

\begin{itemize}
\item \textbf{vgg\_features:} \textit{fc6} layer of VGG19 \cite{simonyan2014very} trained on ImageNet dataset\cite{deng2009imagenet} is used as the feature vector for each image in the snippet. The feature vector for a snippet is an average of all the image feature vectors in that snippet. The feature size is 4096.
\item \textbf{googlenet\_features:} \textit{pool5\_7x7\_s1} layer of GoogLeNet \cite{szegedy2015going} trained on MIT Places dataset \cite{zhou2014learning} is used as the feature vector for each image in the snippet. The feature vector for a snippet is 1024-d and is an average of all the image feature vectors in that snippet.
\item \textbf{vgg\_p\_concepts:} The output probability layer of VGG19 \cite{simonyan2014very} trained on ImageNet dataset \cite{deng2009imagenet} is used as the feature vector for each image in the snippet. The feature vector of a snippet is an average of all the image feature vectors in that snippet. The feature size is 1000.
\item \textbf{googlenet\_p\_concepts:} The output probability layer GoogLeNet \cite{szegedy2015going} trained on MIT Places dataset \cite{zhou2014learning} is used as the feature vector for each image in the snippet. The feature vector of a snippet is an average of all the image feature vectors in that snippet and is 365 dimensional.
\item \textbf{vgg\_concepts:} The output probability layer of VGG19 \cite{simonyan2014very} trained on ImageNet dataset \cite{deng2009imagenet} is used as the feature vector for each image in the snippet. To create the feature vector of a snippet, an average of all the image feature vectors in that snippet is taken. This vector is one-hot-encoded based on a 0.5 threshold. The feature size is 1000.
\item \textbf{goolenet\_concepts:} The output probability layer of GoogLeNet \cite{szegedy2015going} trained on MIT Places dataset \cite{zhou2014learning} is used as the feature vector for each image in the snippet. To create the feature vector of a snippet, an average of all the image feature vectors in that snippet is taken. This vector is one-hot-encoded based on a 0.5 threshold. The feature size is 365.
\item \textbf{yolo\_coco\_concepts:} A vector of size 80 (corresponding to 80 classes in COCO dataset) where each component represents the average number of objects of the respective COCO \cite{lin2014microsoft} class, found by the YOLO \cite{redmon2016you} network trained on COCO dataset, in all images in the snippet.
\item \textbf{yolo\_voc\_concepts:} A vector of size 20 (corresponding to 20 classes in PASCAL VOC dataset) where each component represents the average number of objects of the respective PASCAL VOC\cite{Everingham10} class, found by the YOLO \cite{redmon2016you} network trained on PASCAL VOC dataset, in all images in the snippet.
\item \textbf{yolo\_coco\_p\_concepts:} A vector of size 81 (corresponding to 80 classes in COCO dataset and 1 dummy class) where each component represents the average fraction of objects of the respective COCO \cite{lin2014microsoft} class relative to the total number of objects detected, found by the YOLO \cite{redmon2016you} network trained on COCO dataset, in all images in the snippet. The last component is 1 if no objects were detected.
\item \textbf{yolo\_voc\_p\_concepts:} A vector of size 21 (corresponding to 20 classes in PASCAL VOC dataset and 1 dummy class) where each component represents the average fraction of objects of the respective PASCAL VOC \cite{Everingham10} class relative to the total number of objects detected, found by the YOLO \cite{redmon2016you} network trained on PASCAL VOC dataset, in all images in the snippet. The last component is 1 if no objects were detected.
\item \textbf{color\_hist\_r\_features:} A vector of size 256 representing the average red color histogram of the images in the snippet.
\item \textbf{color\_hist\_g\_features:} A vector of size 256 representing the average green color histogram of the images in the snippet.
\item \textbf{color\_hist\_b\_features:} A vector of size 256 representing the average blue color histogram of the images in the snippet.
\item \textbf{color\_hist\_h\_features:} A vector of size 180 representing the average hue histogram of the images in the snippet.
\item \textbf{color\_hist\_s\_features:} A vector of size 256 representing the average saturation histogram of the images in the snippet.
\end{itemize}}

\subsection{Evaluation measure}

To serve the desired purpose and to be suitable to be used in our framework, we wanted an evaluation measure which would satisfy the following characteristics: 1) The reward for including an $r$ rated snippet must be greater than the reward for including an $r-1$ rated snippet, 2) A negative rated snippet must be penalized, 3) An $r$ rated segment, no matter how big, should not displace an $r+1$ rated segment from a budget constrained gold summary, 4) No number of $r$ rated segments should displace an $r+1$ rated segment from a budget constrained gold summary, 5) In the gold summary, segments marked non-repetitive should not be broken unless it is absolutely necessary (possibly the last one to fit within the boundary) and 6)  No reward should be given for picking more than $\beta$ seconds of a segment marked repetitive. After a careful design involving interactions of different multiplicative and additive terms and their effects on the above characteristics, we came up with the following formulation which satisfies the above characteristics. 

The score function for video V, $S_V: y \rightarrow R$, is defined as:
\begin{align*}
S_V(y) &= \sum\limits_{x_i \in X_P} |y \cap x_i| * (1 + \frac{|y \cap x_i|}{|x_i|}) * e ^ {\alpha * rating(x_i)} \\
&+  \sum\limits_{x_i \in X_R} \min(|y \cap x_i|, \beta) * (1 + \frac{\min(|y \cap x_i|, \beta)}{\min(|x_i|, \beta)}) \\
&* e ^ {\alpha * rating(x_i)} \\
&- \sum\limits_{x_i \in X_N} |y \cap x_i| * k
\end{align*}
where,
$X_P$ is the set of segments in V marked non-repetitive and rated positive, $X_R$ is the set of segments in V marked repetitive and rated positive, $X_R$ is the set of segments in V rated negative, $\alpha > 0$ is the reward scaling hyper-parameter, $\beta > 0$ is the repetitiveness cut-off factor and $k > 0$ is the penalty factor.

This function is neither submodular nor supermodular. However, this can be written as a sum of a submodular and a supermodular function \narxiv{ (details and proof in supplementary material)} and hence the bounds discussed below hold true when this appears as the margin in the discrete optimization of the loss augmented objective (Equation ~\ref{loss-aug}). 
\arxiv{
\begin{lemma}
The evaluation function $S_V(y)$ can be written as a sum of a submodular $f(y)$ and a supermodular function $g(y)$ where,
\begin{align*}
f(y) &= \sum\limits_{x_i \in X_R} \min(|y \cap x_i|, \beta) * e ^ {\alpha * rating(x_i)} \\ &- \sum\limits_{x_i \in X_R} \frac{\max(|y \cap x_i|^2, \beta)}{\min(|x_i|, \beta)} \\ &- \sum\limits_{x_i \in X_N} |y \cap x_i| * k
\end{align*}
and
\begin{align*}
g(y) &= \sum\limits_{x_i \in X_P} |y \cap x_i| * (1 + \frac{|y \cap x_i|}{|x_i|}) * e ^ {\alpha * rating(x_i)} \\ &+ \sum\limits_{x_i \in X_R} \frac{|y \cap x_i|^2}{\min(|x_i|, \beta)}
\end{align*}
\end{lemma}
It is easy to see that the function $f$ and $g$ above are submodular and supermodular respectively, and moreover, expanding the terms we can see that by adding these two functions, we get back $S_V(y)$.
}
\subsection{Discrete Optimization}
Our framework entails two different discrete optimization problems - maximization of the weighted mixture in the loss augmented inference, Equation ~\ref{loss-aug}, and during inference to obtain the summary once the mixture is learnt (Equation ~\ref{max}). For efficient optimization with guaranteed bounds, it is important to understand certain characteristics of these components. Note that our mixture of set functions can be written as follows:
\begin{align*}
f(X) &= 
\alpha f^{\mbox{msub}}(X) + \beta f^{\mbox{nmsub}}(X) + \gamma f^{\mbox{sup}}(X) \\
&+ \delta f^{\mbox{d}}(X)
\end{align*}
where $f^{\mbox{msub}}(X)$ is a monotone submodular function, $f^{\mbox{nmsub}}(X)$ is a non-monotone submodular function, $f^{\mbox{sup}}(X)$ is a monotone supermodular function, and $f^{\mbox{d}}(X)$ is a dispersion function (also called disparity-min) ($f^{\mbox{d}}(X) = \min_{i, j \in X} d_{ij}$), and $\alpha, \beta, \gamma, \delta \geq 0$. Moreover, we assume that each of the functions above are non-negative (without loss of generality). Note that in the above, we have grouped all monotone submodular, non-monotone submodular, supermodular functions together. The only function which is neither submodular nor supermodular is Disparity Min.

We would like to understand the theoretical guarantees for the following optimization problem:
\begin{equation} \label{optprob}
\max \{f(X) \ | \ X \subseteq V, |X| \leq k\}
\end{equation}
for various values of $\alpha, \beta, \gamma, \delta$.
\begin{theorem}
The following theoretical results hold for solving the optimization problem of Equation~\ref{optprob}:
\begin{enumerate}
\item We obtain an approximation factor of $1 - 1/e$ if $\alpha \geq 0$ and $\beta = \gamma = \delta = 0$.
\item We obtain an approximation factor of $1/2$ if $\delta > 0$ and $\alpha = \beta = \gamma = 0$.
\item We obtain an approximation factor of $1/e$ if $\alpha \geq 0, \beta > 0$ and $\gamma = \delta = 0$.
\item We obtain an approximation factor of $1/4$ if $\alpha > 0, \delta > 0$ and $\beta = \gamma = 0$
\item We obtain an approximation factor of $1/2e$ if $\alpha, \beta, \delta \geq 0$ and $\gamma = 0$.
\item We obtain an approximation factor of $\frac{(1 - e^{(1 - \kappa^l)\kappa_k})}{\kappa_k}$ where $k(X) = f^{\mbox{msub}}(X)$ and $l(X) =  f^{\mbox{sup}}(X)$, if $\alpha, \gamma \geq 0$ and $\beta = \delta = 0$.
\item We obtain an approximation factor of $\frac{(1 - e^{(1 - \kappa^l)\kappa_k})}{2\kappa_k}$ where $k(X) = f^{\mbox{msub}}(X)$ and $l(X) =  f^{\mbox{sup}}(X)$, if $\alpha, \gamma, \delta \geq 0$ and $\beta = 0$.
\item The optimization problem of Equation~\ref{optprob} is inapproximable unless P = NP, if $\beta, \gamma > 0$ and $\alpha, \delta \geq 0$.
\end{enumerate}
\end{theorem}
\arxiv{
\begin{proof}
The following paragraphs provide the proofs for each of the cases enumerated above:
\begin{enumerate}
\item The first result follows directly from~\cite{nemhauser1978analysis} since when $\beta = \gamma = \delta = 0$, $f$ is a monotone submodular function. Hence the greedy algorithm admits an approximation guarantee of $1 - 1/e$. 
\item For the second result, notice that when $\alpha = \beta = \gamma = 0$, we have $f(X) = \delta f^{\mbox{d}}(X)$, the dispersion function. Following Lemma 1 from~\cite{dasgupta2013summarization}, the Dispersion function admits a $1/2$ approximation factor for the cardinality constrained optimization problem from Equation~\ref{optprob}.
\item The third result is a consequence of \cite{buchbinder2014submodular}, since we have a non-monotone submodular maximization problem subject to a cardinality constraint. Note that the Randomized Greedy algorithm achieves an approximation guarantee of $1/e$. 
\item The fourth result follows the proof technique from Theorem 1 in~\cite{dasgupta2013summarization}. since when $\alpha, \delta > 0$, $f(X)$ is a sum of a monotone submodular function and a dispersion function. Denote $m(X)$ as the monotone submodular function, and $d(X)$ as the dispersion function. Note that $f(X) = \alpha m(X) + \delta d(X)$. For the algorithm, we first optimize $m(X)$ alone, say using a greedy algorithm and then we optimize $d(X)$ alone, again using the greedy algorithm. We then take the better amongst the solutions achieved. Let $S_m$ denote the solution obtained by optimizing the monotone submodular function, and  $S_d$ denote the solution obtained by optimizing the Dispersion function. Denote $O_m$ and $O_d$ as the optimal solutions respectively. Note that $m(S_m) \geq (1 - 1/e) m(O_m) \geq 1/2 m(O_m)$ and $d(S_d) \geq 1/2 d(O_d)$ from the respective approximation guarantees of the individual algorithms. Denote $X$ as the better amongst $S_m, S_d$ in terms of the final objective $f$. Therefore, $f(X) \geq 1/2 (f(S_d) + f(S_m))$. Note that $f(S_m) \geq \alpha m(S_m)$ and $f(S_d) \geq \delta d(S_d)$ and therefore $f(X) \geq 1/2(\alpha m(S_m) + \delta d(S_d)) \geq 1/4(\alpha m(O_m) + \delta d(O_d)) \geq 1/4 f(O)$ where $O$ is the optimal solution (note that $m(O) \leq m(O_m)$ and $d(O) \leq d(O_d)$ by definition.
\item For the fifth result, we use a similar proof technique as above. Note that we can achieve an approximation factor of $1/e$ by maximizing a non-monotone submodular function subject to a cardinality constraint, and a factor of $1/2$ for maximizing the Dispersion function. Let $n(X)$ as the non-monotone submodular function, and $d(X)$ as the dispersion function. We know that $f(X) = \beta n(X) + \delta d(X)$ (w.l.o.g, take the monotone function term into the non-monotone function since the sum of a monotone and non-monotone function is in general non-monotone). For the algorithm, we first optimize $n(X)$ alone, say using a randomized greedy algorithm and then we optimize $d(X)$ alone, using the greedy algorithm. We then take the better amongst the solutions achieved. Let $S_n$ denote the solution obtained by optimizing the non-monotone submodular function, and  $S_d$ denote the solution obtained by optimizing the Dispersion function. Denote $O_n$ and $O_d$ as the optimal solutions respectively. Note that $n(S_n) \geq 1/e n(O_n)$ and $d(S_d) \geq 1/2 d(O_d) \geq 1/e d(O_d)$ from the respective approximation guarantees of the individual algorithms. Denote $X$ as the better amongst $S_n, S_d$ in terms of the final objective $f$. Therefore, $f(X) \geq 1/2 (f(S_d) + f(S_n))$. Note that $f(S_n) \geq \beta n(S_n)$ and $f(S_d) \geq \delta d(S_d)$ and therefore $f(X) \geq 1/2(\beta n(S_n) + \delta d(S_d)) \geq 1/2e(\alpha n(O_n) + \delta d(O_d)) \geq 1/2e f(O)$ where $O$ is the optimal solution (note that $n(O) \leq n(O_n)$ and $d(O) \leq d(O_d)$ by definition.
\item The sixth result is a direct consequence of~\cite{bai2018greed}, since in this case $f(X)$ is a sum of monotone submodular and a monotone supermodular function. Note that the approximation bounds here depend on the curvature of the two functions.
\item The seventh result follows by combining the proof technique from the fourth and fifth result, with the bound above. In the interest of space, we skip the details of this.
\item Finally, when $\alpha, \beta, \gamma > 0$, we have the most general case of the sum of non monotone submodular and supermodular functions. This is then equivalent to a difference of submodular functions, which is known to be inapproximable, following Theorems 5.1 and 5.2 from~\cite{iyer2012algorithms}.
\end{enumerate}
\end{proof}
}
\narxiv{The proofs for each of the cases enumerated above are included in the extended version of the paper in supplementary material.}

\subsection{Generating Ground Truth Summaries}
The use of ratings allow us to generate multiple ground truth summaries for a video. The total number of possible summaries could be exponential in video duration (a variant of knapsack on duration of segments), so for our experiments we randomly generate up-to 500 ground truth summaries for each video for each budget percentage (5, 15 and 30). Starting from the highest rating, if all segments of that rating can be fit in the budget, they are included in the summary. If all segments of a rating cannot be included in the remaining budget, using a flavor of standard coin exchange problem, maximal combinations of segments are enumerated such that possibly only the last segment gets broken.
\arxiv{
Algorithm~\ref{alg:gtGen} demonstrates how we generate the ground truth summaries given a set of ratings and a budget.
\begin{algorithm}
\caption{Generate ground truth summaries of a video for a given $budget$}
\label{alg:gtGen}
\begin{algorithmic}
\Procedure{generateGTSummaries}{$segments$, $budget$}
\State $gtSet = \{\}$
\State $normalizedSegments = \{\}$
\State $remBudget = budget$
\For {$s$ in $segments$}
\If {$s.repetitive == true$}
   	\State $s' = s$ trimmed by $\beta$
   	\State $normalizedSegments.add(s')$
   \Else
   	\State $normalizedSegments.add(s)$
   \EndIf
\EndFor
\State $tempGt = \{\}$
\For {$r$ in $3,2,1,0$}
\State $currSegs =$ segments rated $r$ in $normalizedSegments$
\State $cumBudget =$ Total duration of $currSegs$
   \If {$cumBudget < remBudget$}
   	\State $tempGt.addSnippets(currSegs)$
       \State $remBudget -= currBudget$
   \Else\If {$cumBudget == remBudget$}
   	\State $tempGt.addSnippets(currSegs)$
       \State $gtSet.add(tempGt)$
       \State \textbf{break}
    \Else
    \While {$i <= 500$}
       	\State $rndSegSet$ = $X \subset currSegs$ and $\sum\limits_{x \in X} x.duration = remBudget$
           \State $gtSet.add(tempGt \cup toSnippets(rndSegs))$
       	\State $i++$
       \EndWhile
    \EndIf
   \EndIf
\EndFor
\Return $gtSet$
\EndProcedure
\end{algorithmic}
\end{algorithm}
}

\section{Experiments and Results}
What follows is a description of various experiments performed and results observed using the above framework for domain specific video summarization. For surveillance videos (entry exit and office), since night videos are black and white we do not use the color histogram features based on hue and saturation. For videos in Soccer, Cricket and Birthday domains we additionally perform shot detection to identify distinct shots in the video and create a feature which keeps track of the snippets present in each shot. This is consumed by the continuity component in the mixture. However, we do not use this continuity component in the mixture for the surveillance videos where the notion of a shot is not well defined in those videos. Also unless explicitly stated, for training, during each epoch, a random ground truth summary, out of the many possible summaries, was chosen for each video so that over a large number of epochs, all ground truths get covered. We do a train test split of 70-30 with respect to the number of videos in each domain in the dataset. The hyper-parameters used for the evaluation measure while training were $\alpha = 1$, $\beta = 6$ and $k=2$. We arrive at best values for $\lambda_1$ and $\lambda_2$ by testing the models on the held-out validation set.

\paragraph{Sanity of ground truths and behavior of evaluation measure}
We perform the following sanity checks on the ground truth summaries produced and the behavior of our evaluation measure:
\begin{itemize}
\item scores of all ground truth summaries of a particular budget for a video should be same, asserting that the synthesis of ground truths as above is consistent with $S_V(y)$
\item scores of ground truth summaries should always be greater than randomly produced summaries - for all lengths, for all videos for all categories
\end{itemize}

In this experiment we compare the normalized scores of the ground truth summaries against the normalized scores of 1000 random summaries picked exclusively from segments rated highly positive. We do the standard min-max normalization. We plot the minimum, maximum and average scores for both the random summaries and the ground truth summaries. To ensure that the random summaries do not get very low scores (and hence favoring the ground truth summaries during comparison), the random summaries used in this experiment are not truly random. They do not include the negatively rated segments.

\arxiv{\begin{figure}
\centering
  \begin{tabular}{@{}cccc@{}}
    \includegraphics[width=.25\textwidth]{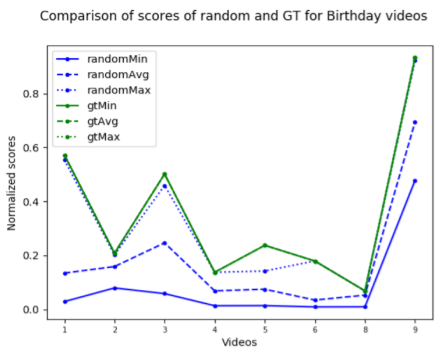} &
    \includegraphics[width=.25\textwidth]{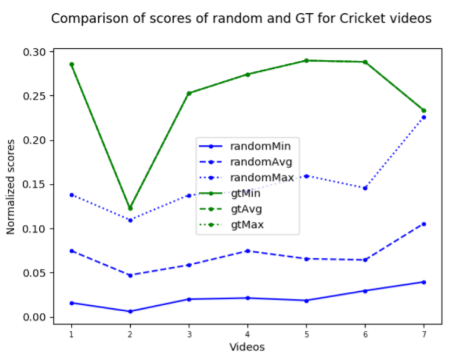} \\
    \includegraphics[width=.25\textwidth]{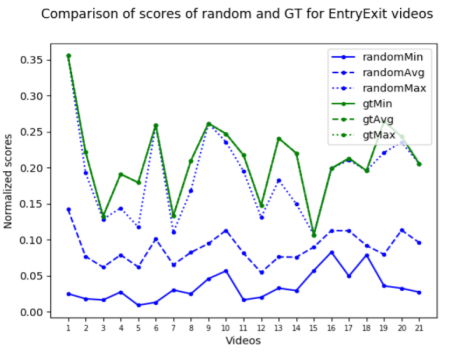} &
    \includegraphics[width=.25\textwidth]{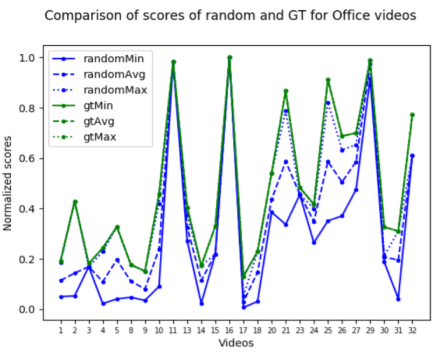} \\
    \multicolumn{2}{c}{\includegraphics[width=.25\textwidth]{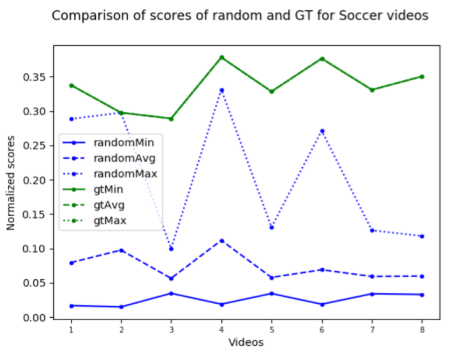}}
  \end{tabular}
  \caption{Scores of ground truth summaries compared with random summaries}
  \label{pic:rndvsGt}
\end{figure}}

\arxiv{The results as per figure \ref{pic:rndvsGt} show that our evaluation measure and our ground truth generation algorithm behave as expected. All ground truth summaries have the same normalized scores and hence the minimum, maximum and the average scores for all ground truth summaries coincide as one line in the plots. Also all ground truth summaries score higher than any random summary across all domains and all videos. }\narxiv{The results (included in the extended version) show that all ground truth summaries score higher than any random summary across all domains and all videos.} 
We also verify the ground truth summaries visually by representing  them as videos and visually assessing their quality.

\paragraph{Learning experiments}
For learning weights in the above formulation, we constrain the weights of all submodular components to be always positive. There is no such constraint for modular components in the mixture. We compare AdaGrad and stochastic gradient descent and find AdaGrad to work better for all experiments and hence all results reported use AdaGrad. The reported numbers are losses (i.e. $normalizedGTScore$ - $normalizedSummaryScore$). We compare the following results: a) \emph{All Modular: } Train with only modular terms in the mixture, b) \emph{All Submodular: } Train with only submodular terms in the mixture, c) \emph{Full: } Complete Mixture (all modular terms and all submodular terms).

We compare all these to random summaries, uniform summaries and the best individual component baselines instantiated with different features are used individually to produce summaries. Results are reported in Table \ref{tab:learningResults}. We observe that combining both the submodular and modular terms in the mixture (full) provides the best results, as compared to just using submodular and modular terms alone or as compared to any of the baselines. Moreover, the learnt mixtures for only modular and either submodular also outperform Random, Uniform and the average individual submodular functions. We see that the best individual submodular functions also perform better than random or uniform baselines, but not as well as the learnt mixtures, thus proving the benefit of learning for this problem. We also verify the goodness of a summary both quantitatively (using the scores from our evaluation measure) and qualitatively (by visualizing the summary produced). This establishes our hypothesis that joint training can significantly help. 

\begin{table}[ht]
\begin{center}
\begin{tabular}{|l|l|r||}
\hline
\textbf{Domain} & \textbf{Method} & \textbf{ScoreLoss} \\
\hline
\multirow{6}{*}{Birthday} & All Modular & 0.7234 \\ \cline{2-3}
						& All Submodular &  0.7307 \\ \cline{2-3}
                        & Full & \textbf{0.6625} \\ \cline{2-3}
                        & Random & 0.7378 \\ \cline{2-3}
                        & Uniform & 0.7569 \\ \cline{2-3}
                        & Submodular & 0.7432 \\ \cline{2-3}
\hline

\multirow{6}{*}{EntryExit} & All Modular & 0.5967 \\ \cline{2-3}
						& All Submodular &  0.6306 \\ \cline{2-3}
                        & Full & \textbf{0.5884} \\ \cline{2-3}
                        & Random & 0.7706 \\ \cline{2-3}
                        & Uniform & 0.7785 \\ \cline{2-3}
                        & Submodular & 0.6306 \\ \cline{2-3}
\hline

\multirow{6}{*}{Cricket} & All Modular & 0.8140 \\ \cline{2-3}
						& All Submodular &  0.8275 \\ \cline{2-3}
                        & Full & \textbf{0.7733} \\ \cline{2-3}
                        & Random & 0.8911 \\ \cline{2-3}
                        & Uniform & 0.8979 \\ \cline{2-3}
                        & Submodular & 0.8275 \\ \cline{2-3}
                        
\hline

\multirow{6}{*}{Office} & All Modular & 0.3871 \\ \cline{2-3}
						& All Submodular &  0.4783 \\ \cline{2-3}
                        & Full & \textbf{0.3696} \\ \cline{2-3}
                        & Random & 0.5743 \\ \cline{2-3}
                        & Uniform & 0.5399 \\ \cline{2-3}
                        & Submodular & 0.5590 \\ \cline{2-3}
                        
\hline

\multirow{6}{*}{Soccer} & All Modular & 0.8849 \\ \cline{2-3}
						& All Submodular &  0.7645 \\ \cline{2-3}
                        & Full & \textbf{0.6533} \\ \cline{2-3}
                        & Random & 0.9217 \\ \cline{2-3}
                        & Uniform & 0.8747 \\ \cline{2-3}
                        & Submodular & 0.9152 \\ \cline{2-3}
                        
\hline
\end{tabular}
\end{center}
\caption{Learning experiments comparing all submodular, all modular, full mixture with random, uniform and submodular baselines for all domains}
\label{tab:learningResults}
\end{table}

\paragraph{Verification of learning domain specific characteristics}

To demonstrate that the learnt summaries are domain specific, we test the model learnt on one domain in producing summaries of another domain. 
The results in Table \ref{tab:domSpec} shows that models learnt on one domain perform poorly on other domains, establishing that the model has indeed learnt domain specific characteristics.

\begin{table}[ht]
\begin{center}
\begin{tabular}{|l|l|r|}
\hline
\textbf{Model Trained On} & \textbf{Model Tested On} & \textbf{ScoreLoss} \\
\hline
\multirow{3}{*}{Birthday} & Birthday & \textbf{0.6625} \\ \cline{2-3}
						& Soccer & 0.9753  \\ \cline{2-3}
                        & Cricket & 0.9177 \\ \cline{2-3}
\hline
\multirow{4}{*}{EntryExit} & EntryExit & \textbf{0.5884} \\ \cline{2-3}
						& Soccer & 0.9900  \\ \cline{2-3}
                        & Cricket & 0.9710 \\ \cline{2-3}
                        & Birthday & 0.8009 \\ \cline{2-3}
\hline
\multirow{3}{*}{Cricket} & Cricket & \textbf{0.7733} \\ \cline{2-3}
						& Soccer & 0.8284  \\ \cline{2-3}
						& Birthday & 0.8103 \\ \cline{2-3}
\hline
\end{tabular}
\end{center}
\caption{Sample results for cross domain experiments}
\label{tab:domSpec}
\end{table}

\begin{figure}[!htbp]
\begin{center}
\includegraphics[scale=0.25]{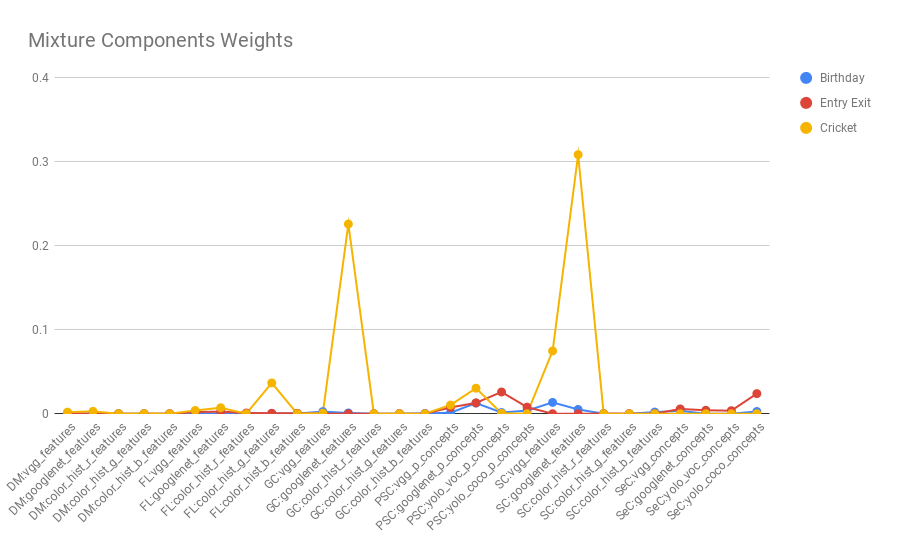}
\includegraphics[scale=0.2]{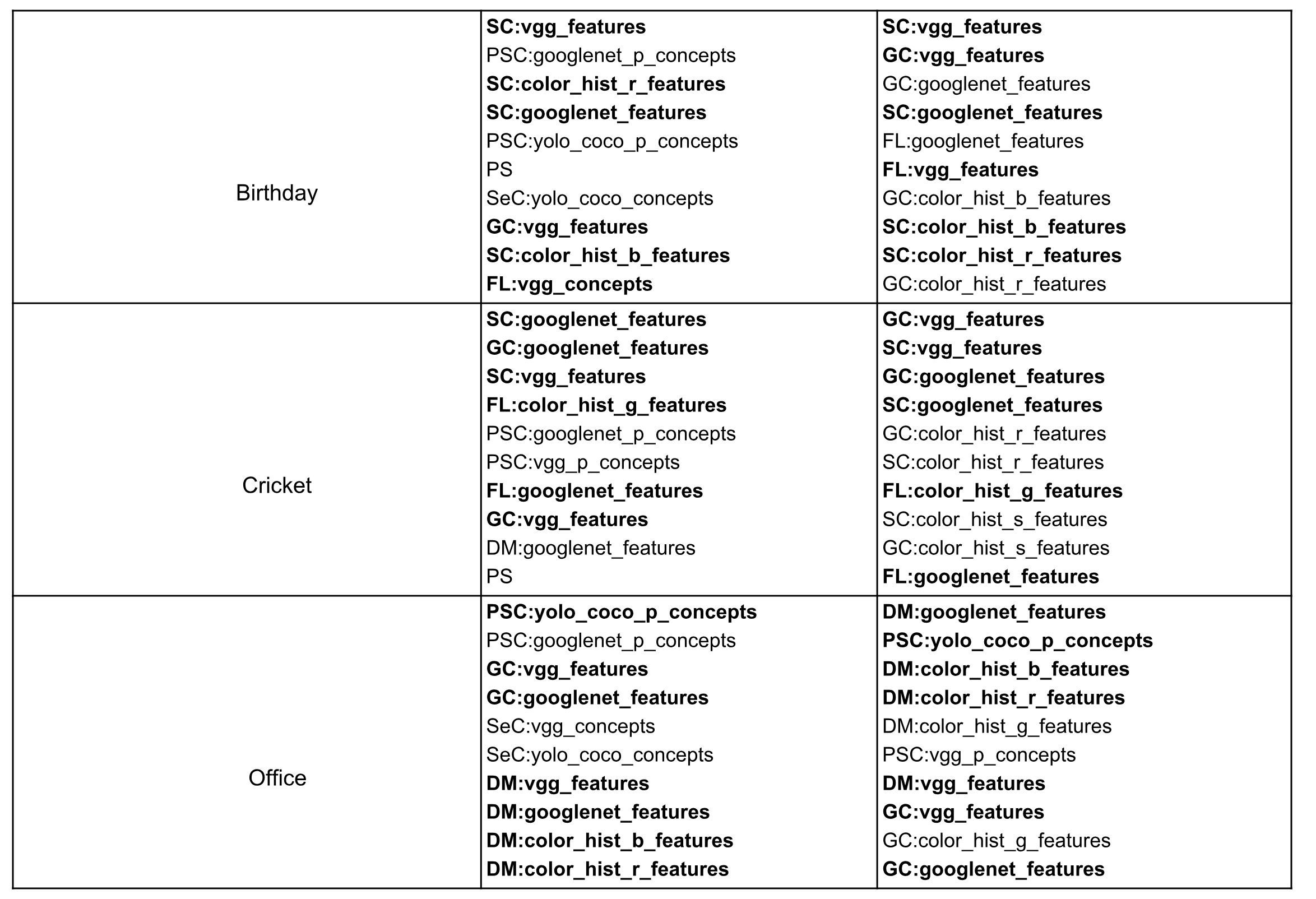}
\caption{Sample results for a few domains: (top) Mixture component weights and (bottom) the top components based on mixture weights and top components when evaluated in isolation.}
\label{pic:compWeights}
\end{center}
\end{figure}

Next, we look at the weights learnt for the different submodular components. Figure \ref{pic:compWeights} (top) shows magnitude of the learnt weights the  different domains. We see that different domains prefer different submodular components and features to produce good summaries. For example, scene features (googlenet\_features) are important for Cricket, while object detection features (yolo\_voc\_p\_concepts and yolo\_coco\_concepts) are more important for Surveillance Videos. For Cricket, the scene of ground or pitch or crowd has a lot of bearing on the importance and for Surveillance, detection of entities assume more importance, given the static scene. Next, we look at the correlation between the components which achieve the best weight in the learnt mixture and the components which achieve the best score when run in isolation. We see in the bottom table of Figure~\ref{pic:compWeights} that there is a strong correlation between the two. In particular about 6 to 7 out of the top ten components are the same in both buckets. It is also informative to look at the components themselves. For Birthday and Cricket, we see that Saturated Coverage and Facility Location (i.e. the representative models) are the winners, while in Office (which are surveillance videos), we see a lot of Disparity Min (diversity) functions as winners. \narxiv{A more detailed view of weights learnt for different domains as well as tests demonstrating significance of our results are provided in the extended version of this paper in the supplementary material.}

\arxiv{The Figures \ref{pic:birthdayWeights}, \ref{pic:cricketWeights} and \ref{pic:entryExitWeights} show the different weights learnt for different domains.

\begin{figure}[!htbp]
\begin{center}
\includegraphics[scale=0.5]{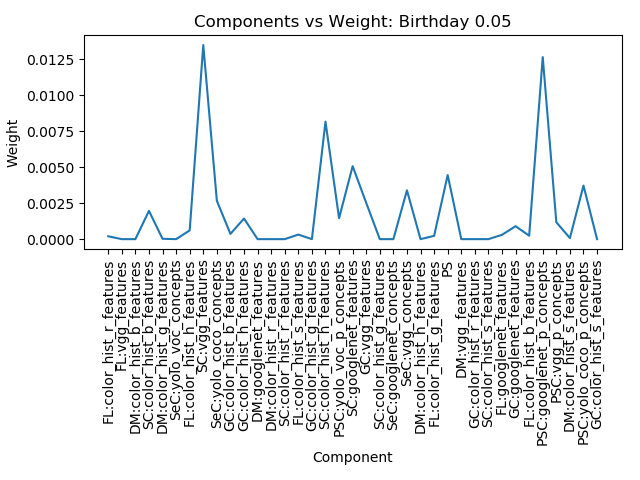}
\caption{Birthday Weights}
\label{pic:birthdayWeights}
\end{center}
\end{figure}

\begin{figure}[!htbp]
\begin{center}
\includegraphics[scale=0.5]{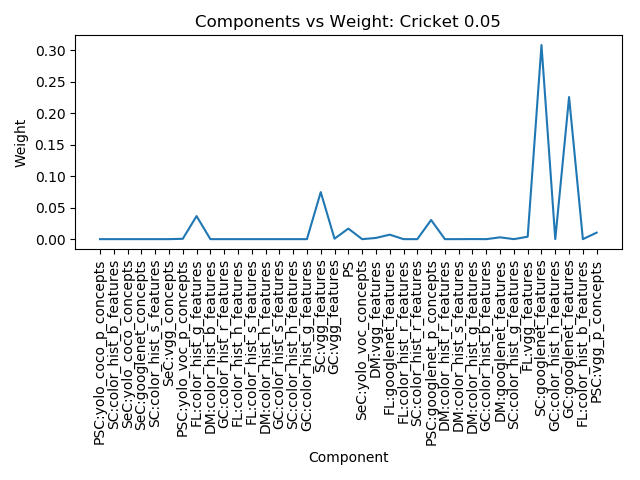}
\caption{Cricket Weights}
\label{pic:cricketWeights}
\end{center}
\end{figure}

\begin{figure}[!htbp]
\begin{center}
\includegraphics[scale=0.5]{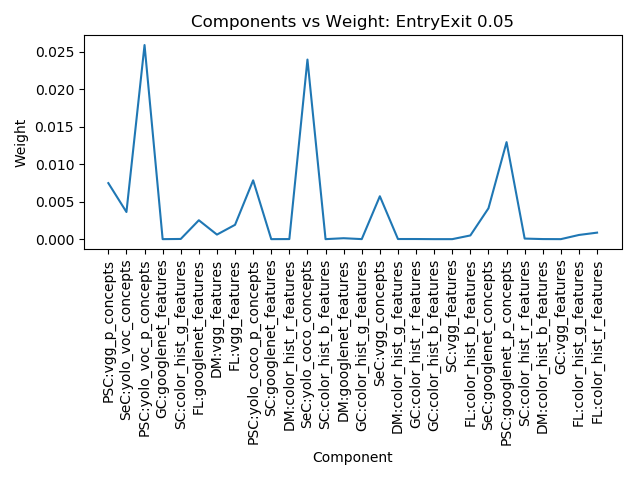}
\caption{EntryExit Weights}
\label{pic:entryExitWeights}
\end{center}
\end{figure}
}

\begin{figure}[!htbp]
\begin{center}
\includegraphics[scale=0.18]{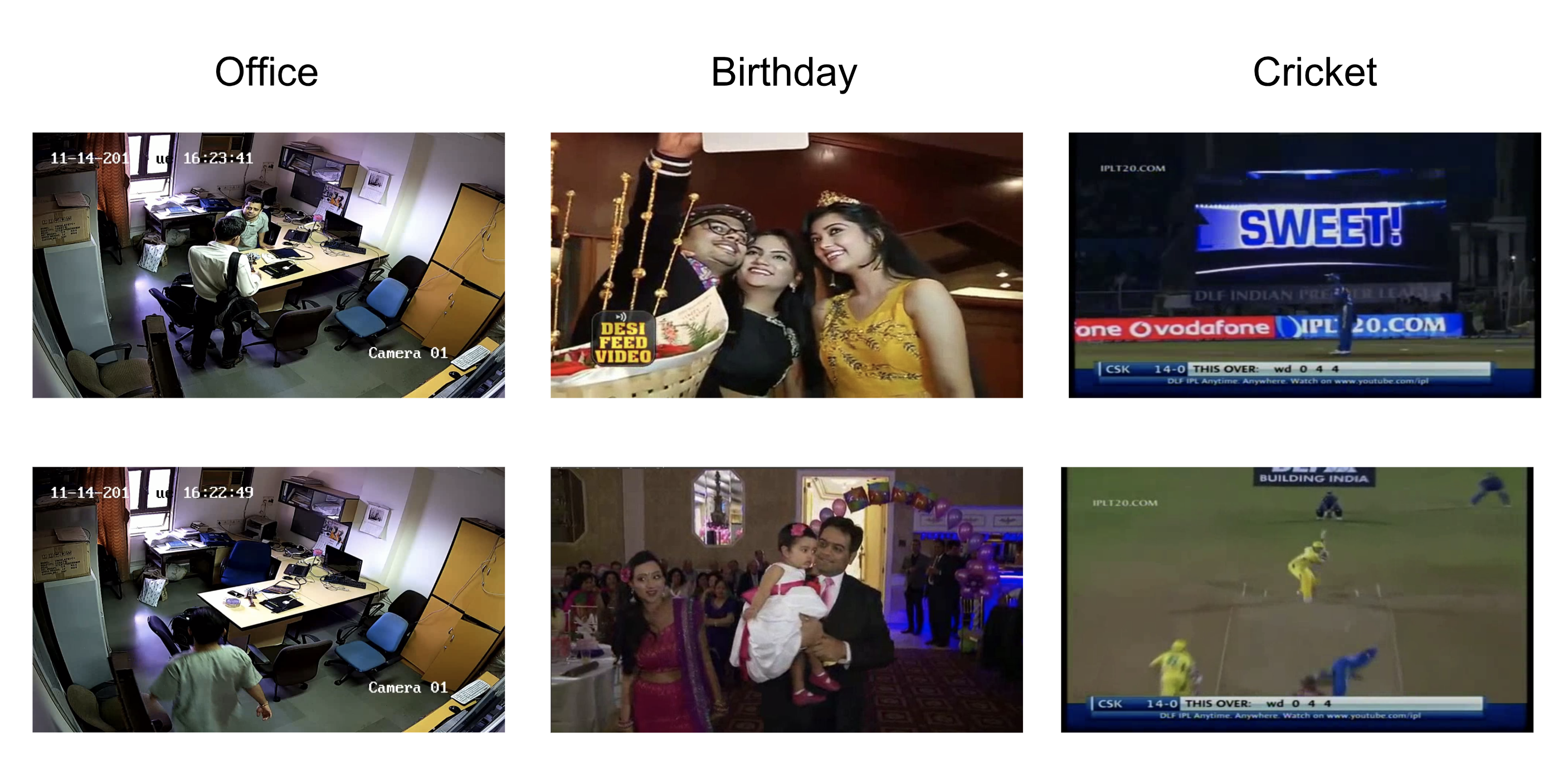}
\caption{Top frames based on the domain specific learnt mixtures for Office, Birthday and Cricket}
\label{pic:topFrames}
\end{center}
\end{figure}

Finally, we look at the top ranked frames based on the learnt mixture for each domain. We see that the the frames intuitively capture the most important aspects for that domain. For example, in office footages, the top frames are frames where people are either entering/leaving or meeting, in Birthday videos, we see a shot where people are taking a selfie and a shot where the birthday girl is posing are ranked the best. Similarly in Cricket, shots where the player is about to hit, and where there is a four being hit are selected. Hence we see that the joint training has learnt specific domain specific importance of events along with the right weights for diversity and representation. 

%We also verify that the weights learnt for the modular terms are in good correlation with what is important for that domain by visualizing those snippets (which get high values for those modular features) to see that they match with scenes which are more important for a domain. In addition, we use t-SNE visualization of the top ranked modular terms (feature type) to cluster the snippets of the videos and find that all important snippets fall in the same cluster. More details on the above has been provided in the supplementary material.

\paragraph{Importance of ratings in generating multiple ground truths}

To show that ability to generate multiple ground truth from our ratings based annotation is beneficial, we compare the scores obtained by training with a single ground truth summary in every epoch against the scores obtained by training with a random ground truth summary in every epoch.

\begin{table}[ht]
\begin{center}
\scalebox{0.7}{
\begin{tabular}{|l|l|r|}
%\textbf{Domain} & \textbf{Method} & \textbf{ScoreLoss} \\
\hline
\multirow{2}{*}{Birthday} & Random GTs & \textbf{0.6625} \\ \cline{2-3}
						& Same GT & 0.6818 \\
\hline
\multirow{2}{*} {EntryExit} & Random GTs & \textbf{0.5883} \\ \cline{2-3}
						 & Same GT & 0.6188  \\
\hline
\end{tabular}
}
\end{center}
\caption{Sample results for using multiple ground truths}
\label{tab:rndGtVsfixedGt}
\end{table}

The results on Table \ref{tab:rndGtVsfixedGt} suggests that the model indeed learns better when multiple ground truth summaries are used hence establishing the importance of the system of ratings which allow us to generate many ground truths.

\section{Conclusion}
Motivated by the fact that what makes a good summary differs across domains, we set out to develop a framework which would automatically learn what is considered important for a domain, both in terms of the kind of snippets to be selected and also in terms of the desired characteristics of the summary produced in terms of representativeness, coverage, diversity, {\em etc.} We also establish that ratings provide a more efficient way of supervision to impart domain knowledge necessary to create such summaries. Further, we propose a novel evaluation measure well suited for this task. In the absence of any existing dataset which would lend itself well to this particular problem, we created a gold standard dataset and will be making it public as a part of this work. Through several experiments we demonstrated the effectiveness of our solution in producing domain specific summaries which can be seen as a first significant breakthrough in this direction.

\bibliography{main}
\bibliographystyle{ieee}

\end{document}